\documentclass[10pt,twocolumn,letterpaper]{article}

\usepackage{cvpr}
\usepackage{times}
\usepackage{epsfig}
\usepackage{graphicx}
\usepackage{amsmath}
\usepackage{amssymb}

\usepackage{bm}
\graphicspath{{Figures/}}
\usepackage{subfigure}
\newcommand{\matr}[1]{\mathbf{#1}}

\usepackage[flushleft]{threeparttable}

\usepackage[pagebackref=true,breaklinks=true,letterpaper=true,colorlinks,bookmarks=false]{hyperref}

 \cvprfinalcopy 

\def\cvprPaperID{3436} 

\ifcvprfinal\fi
\begin{document}

\title{Independently Recurrent Neural Network (IndRNN): Building A Longer and Deeper RNN}

\author{
  Shuai Li \textsuperscript{*}, Wanqing Li \textsuperscript{*}, Chris Cook \textsuperscript{*}, Ce Zhu \textsuperscript{\dag}, Yanbo Gao \textsuperscript{\dag}\\
\textsuperscript{*}School of Computing and Information Technology, University of Wollongong\\
\textsuperscript{\dag}School of Electronic Engineering, University of Electronic Science and Technology of China\\
  \tt\small {\{sl669,wanqing,ccook\}@uow.edu.au,eczhu@uestc.edu.cn,yanbogao@std.uestc.edu.cn}
}

\maketitle

\begin{abstract}
Recurrent neural networks (RNNs) have been widely used for processing sequential data. However, RNNs are commonly difficult to train due to the well-known gradient vanishing and exploding problems and hard to learn long-term patterns. Long short-term memory (LSTM) and gated recurrent unit (GRU) were developed to address these problems, but the use of hyperbolic tangent and the sigmoid action functions results in gradient decay over layers. Consequently, construction of an efficiently trainable deep network is challenging. In addition, all the neurons in an RNN layer are entangled together and their behaviour is hard to interpret. To address these problems, a new type of RNN, referred to as independently recurrent neural network (IndRNN), is proposed in this paper, where neurons in the same layer are independent of each other and they are connected across layers. We have shown that an IndRNN can be easily regulated to prevent the gradient exploding and vanishing problems while allowing the network to learn long-term dependencies. Moreover, an IndRNN can work with non-saturated activation functions such as relu (rectified linear unit) and be still trained robustly. Multiple IndRNNs can be stacked to construct a network that is deeper than the existing RNNs. Experimental results have shown that the proposed IndRNN is able to process very long sequences (over $5000$ time steps), can be used to construct very deep networks ($21$ layers used in the experiment) and still be trained robustly. Better performances have been achieved on various tasks by using IndRNNs compared with the traditional RNN and LSTM. {\color{red}{The code is available at \url{https://github.com/Sunnydreamrain/IndRNN_Theano_Lasagne.}}}
\end{abstract}

\section{Introduction}

Recurrent neural networks (RNNs) \cite{jordan1997serial} have been widely used in sequence learning problems such as action recognition \cite{donahue2015long}, scene labelling \cite{byeon2015scene} and language processing \cite{cho2014learning}, and have achieved impressive results. Compared with the feed-forward networks such as the convolutional neural networks (CNNs), a RNN has a recurrent connection where the last hidden state is an input to the next state. The update of states can be described as follows:
\begin{equation}
\mathbf{h}_t=\sigma(\mathbf{Wx}_t+\mathbf{Uh}_{t-1}+\mathbf{b})
\label{RNN}
\end{equation}
where $\mathbf{x}_t\in \mathbb{R}^{M}$ and $\mathbf{h}_t\in \mathbb{R}^{N}$ are the input and hidden state at time step $t$, respectively. $\mathbf{W}\in \mathbb{R}^{N\times M}$, $\mathbf{U}\in \mathbb{R}^{N\times N}$ and $\mathbf{b}\in \mathbb{R}^{N}$ are the weights for the current input and the recurrent input, and the bias of the neurons. $\sigma$ is an element-wise activation function of the neurons, and $N$ is the number of neurons in this RNN layer.

Training of the RNNs suffers from the gradient vanishing and exploding problem due to the repeated multiplication of the recurrent weight matrix. Several RNN variants such as the long short-term memory (LSTM) \cite{greff2017lstm,jozefowicz2015empirical} and the gated recurrent unit (GRU) \cite{cho2014learning} have been proposed to address the gradient problems. However, the use of the hyperbolic tangent and the sigmoid functions as the activation function in these variants results in gradient decay over layers. Consequently, construction and training of a deep LSTM or GRU based RNN network is practically difficult. By contrast, existing CNNs using non-saturated activation function such as relu can be stacked into a very deep network (e.g. over 20 layers using the basic convolutional layers and over 100 layers with residual connections \cite{he2016deep}) and be still trained efficiently. Although residual connections have been attempted for LSTM models in several works \cite{wu2016google,pradhanexploring}, there have been no significant improvement (mostly due to the reason that gradient decays in LSTM with the use of the hyperbolic tangent and the sigmoid functions as mentioned above).

Moreover, the existing RNN models share the same component $\sigma(\mathbf{Wx}_t+\mathbf{Uh}_{t-1}+\mathbf{b})$ in (\ref{RNN}), where the recurrent connection entangles all the neurons. This makes it hard to interpret and understand the roles of the trained neurons (e.g., what patterns each neuron responds to) since the simple visualization of the outputs of individual neurons~\cite{karpathy2015visualizing} is hard to ascertain the function of one neuron without considering the others.

In this paper, a new type of RNN, referred to as independently recurrent neural network (IndRNN), is proposed. In the proposed IndRNN, the recurrent inputs are processed with the Hadamard product as $\mathbf{h}_t=\sigma(\mathbf{Wx}_t+\mathbf{u}\odot\mathbf{h}_{t-1}+\mathbf{b})$. This provides a number of advantages over the traditional RNN including:

\begin{itemize}
\item The gradient backpropagation through time can be regulated to effectively address the gradient vanishing and exploding problems.
\item Long-term memory can be kept with IndRNNs to process long sequences. Experiments have demonstrated that an IndRNN can well process sequences over $5000$ steps while LSTM could only process less than $1000$ steps.
\item An IndRNN can work well with non-saturated function such as relu as activation function and be trained robustly. 
\item Multiple layers of IndRNNs can be efficiently stacked, especially with residual connections over layers, to increase the depth of the network. An example of 21 layer-IndRNN is demonstrated in the experiments for language modelling.
\item Behaviour of IndRNN neurons in each layer are easy to interpret due to the independence of neurons in each layer.
\end{itemize}
Experiments have demonstrated that IndRNN performs much better than the traditional RNN and LSTM models on the tasks of the adding problem, sequential MNIST classification, language modelling and action recognition.

\vspace{0.5cm}
\section{Related Work}
To address the gradient exploding and vanishing problems in RNNs, variants of RNNs have been proposed and typical ones are the long short-term memory (LSTM) \cite{hochreiter1997long}, and the gated recurrent unit (GRU) \cite{cho2014learning}. Both LSTM and GRU enforce a constant error flow over time steps and use gates on the input and the recurrent input to regulate the information flow through the network. However, the use of gates makes the computation not parallelable and thus increases the computational complexity of the whole network. To process the states of the network over time in parallel, the recurrent connections are fixed in \cite{bradbury2016quasi,lei2017training}. While this strategy greatly simplifies the computational complexity, it reduces the capability of their RNNs since the recurrent connections are no longer trainable. In \cite{arjovsky2015unitary, wisdom2016full}, a unitary evolution RNN was proposed where the unitary recurrent weights are defined empirically. In this case, the norm of the backpropagated gradient can be bounded without exploding. By contrast, the proposed IndRNN solves the gradient exploding and vanishing problems without losing the power of trainable recurrent connections and without involving gate parameters. 

In addition to changing the form of the recurrent neurons, works on initialization and training techniques, such as initializing the recurrent weights to a proper range or regulating the norm of the gradients over time, were also reported in addressing the gradient problems. In \cite{le2015simple}, an initialization technique was proposed for an RNN with relu activation, termed as IRNN, which initializes the recurrent weight matrix to be the identity matrix and bias to be zero. In \cite{talathi2015improving}, the recurrent weight matrix was further suggested to be a positive definite matrix with the highest eigenvalue of unity and all the remainder eigenvalues less than 1. In \cite{neyshabur2016path}, the geometry of RNNs was investigated and a path-normalized optimization method for training was proposed for RNNs with relu activation. In \cite{krueger2016regularizing}, a penalty term on the squared distance between successive hidden states' norms was proposed to prevent the exponential growth of IRNN's activation. Although these methods help ease the gradient exploding, they are not able to completely avoid the problem (the eigenvalues of the recurrent weight matrix may still be larger than 1 in the process of training). Moreover, the training of an IRNN is very sensitive to the learning rate. When the learning rate is large, the gradient is likely to explode. The proposed IndRNN solves gradient problems by making the neurons independent and constraining the recurrent weights. It can work with relu and be trained robustly. As a result, an IndRNN is able to process very long sequences (e.g. over $5000$ steps as demonstrated in the experiments).

On the other hand, comparing with the deep CNN architectures which could be over 100 layers such as the residual CNN \cite{he2016deep} and the pseudo-3D residual CNN (P3D) \cite{qiu2017learning}, most of the existing RNN architectures only consist of several layers (2 or 3 for example \cite{krueger2016zoneout,shahroudy2016ntu,le2015simple}). This is mostly due to the gradient vanishing and exploding problems which result in the difficulty in training a deep RNN. Since all the gate functions, input and output modulations in LSTM employ sigmoid or hyperbolic tangent functions as the activation function, it suffers from the gradient vanishing problem over layers when multiple LSTM layers are stacked into a deep model. Currently, a few models were reported that employ residual connections \cite{he2016deep} between LSTM layers to make the network deeper \cite{wu2016google}. However, as shown in \cite{pradhanexploring}, the deep LSTM model with the residual connections does not efficiently improve the performance. This may be partly due to the gradient decay over LSTM layers. On the contrary, for each time step, the proposed IndRNN with relu works in a similar way as CNN. Multiple layers of IndRNNs can be stacked and be efficiently combined with residual connections, leading to a deep RNN.
\vspace{0.2cm}

\section{Independently Recurrent Neural Network}
In this paper, we propose an independently recurrent neural network (IndRNN). It can be described as:
\begin{equation}
\mathbf{h}_t=\sigma(\mathbf{Wx}_t+\mathbf{u}\odot\mathbf{h}_{t-1}+\mathbf{b})
\label{IndRNN}
\end{equation}
where recurrent weight $\mathbf{u}$ is a vector and $\odot$ represents Hadamard product. Each neuron in one layer is independent from others and connection between neurons can be achieved by stacking two or more layers of IndRNNs as presented later. For the $n$-th neuron, the hidden state $h_{n,t}$ can be obtained as 
\begin{equation}
h_{n,t}=\sigma(\mathbf{w}_{n}\mathbf{x}_t+u_nh_{n,t-1}+b_n)
\label{kernel_form}
\end{equation}
where $\mathbf{w}_{n}$ and $u_n$ are the $n$-th row of the input weight and recurrent weight, respectively. Each neuron only receives information from the input and its own hidden state at the previous time step. That is, each neuron in an IndRNN deals with one type of spatial-temporal pattern independently. Conventionally, a RNN is treated as multiple layer perceptrons over time where the parameters are shared. Different from the conventional RNNs, the proposed IndRNN provides a new perspective of recurrent neural networks as independently aggregating spatial patterns (i.e. through $w$) over time (i.e. through $u$). The correlation among different neurons can be exploited by stacking two or multiple layers. In this case, each neuron in the next layer processes the outputs of all the neurons in the previous layer.

The gradient backpropagation through time for an IndRNN and how it addresses the gradient vanishing and exploding problems are described in the next Subsection \ref{BPTT}. Details on the exploration of cross-channel information are explained in Subsection \ref{relationRNN}. Different deeper and longer IndRNN network architectures are discussed in Subsection \ref{rnn_arcs}.

\vspace{0.3cm}
\subsection{Backpropagation Through Time for An IndRNN}
\label{BPTT}
For the gradient backpropagation through time in each layer, the gradients of an IndRNN can be calculated independently for each neuron since there are no interactions among them in one layer. For the $n$-th neuron $h_{n,t}=\sigma(\mathbf{w}_{n}\mathbf{x}_t+u_nh_{n,t-1})$ where the bias is ignored, suppose the objective trying to minimize at time step $T$ is $J_n$. Then the gradient back propagated to the time step $t$ is 
\begin{small}
{
\begin{align}
\frac{\partial {J_n}}{\partial {h}_{n,t}} &= \
\frac{\partial {J_n}}{\partial {h}_{n,T}}\frac{\partial {h}_{n,T}}{\partial {h}_{n,t}}  
=\frac{\partial {J_n}}{\partial {h}_{n,T}}\
\prod_{k=t}^{T-1} \frac{\partial {h}_{n,k+1}}{\partial {h}_{n,k}} \nonumber \\
&=\frac{\partial {J_n}}{\partial {h}_{n,T}} \prod_{k=t}^{T-1} {\sigma'}_{n,k+1} {u}_n \ 
=\frac{\partial {J_n}}{\partial {h}_{n,T}} {u}_n^{T-t} \prod_{k=t}^{T-1} {\sigma'}_{n,k+1} \
\label{KIRNNgradient}
\end{align}
}
\end{small}
where ${\sigma'}_{n,k+1}$ is the derivative of the element-wise activation function. It can be seen that the gradient only involves the exponential term of a scalar value $u_n$ which can be easily regulated, and the gradient of the activation function which is often bounded in a certain range. Compared with the gradients of an RNN ($\frac{\partial J}{\partial h_T} \prod_{k=t}^{T-1} diag(\sigma'(h_{k+1})) \matr{U}^T $ where $diag(\sigma'(h_{k+1}))$ is the Jacobian matrix of the element-wise activation function), the gradient of an IndRNN directly depends on the value of the recurrent weight (which is changed by a small magnitude according to the learning rate) instead of matrix product (which is mainly determined by its eigenvalues and can be changed significantly even though the change to each matrix entries is small \cite{parlett1964laguerre}). Thus the training of an IndRNN is more robust than a traditional RNN. To solve the gradient exploding and vanishing problem over time, we only need to regulate the exponential term ``${u}_n^{T-t} \prod_{k=t}^{T-1} {\sigma'}_{n,k+1}$'' to an appropriate range. This is further explained in the following together with keeping long and short memory in an IndRNN.

To keep long-term memory in a network, the current state (at time step $t$) would still be able to effectively influence the future state (at time step $T$) after a large time interval. Consequently, the gradient at time step $T$ can be effectively propagated to the time step $t$. By assuming that the minimum effective gradient is $\epsilon$, a range for the recurrent weight of an IndRNN neuron in order to keep long-term memory can be obtained. Specifically, to keep a memory of $T-t$ time steps, $|u_n|\in [\sqrt[\leftroot{-3}\uproot{5}(T-t)]{\frac{\epsilon}{\prod_{k=t}^{T-1} {\sigma'}_{n,k+1}}}, +\infty)$ according to (\ref{KIRNNgradient}) (ignoring the gradient backpropagated from the objective at time step $T$). That is, to avoid the gradient vanishing for a neuron, the above constraint should be met. In order to avoid the gradient exploding problem, the range needs to be further constrained to $|u_n|\in [\sqrt[\leftroot{-3}\uproot{5}(T-t)]{\frac{\epsilon}{\prod_{k=t}^{T-1} {\sigma'}_{n,k+1}}}, \sqrt[\leftroot{-3}\uproot{5}(T-t)]{\frac{\gamma}{\prod_{k=t}^{T-1} {\sigma'}_{n,k+1}}}]$ where $\gamma$ is the largest gradient value without exploding. For the commonly used activation functions such as relu and tanh, their derivatives are no larger than $1$, i.e., $|{\sigma'}_{n,k+1}|\leq 1$. Especially for relu, its gradient is either $0$ or $1$. Considering that the short-term memories can be important for the performance of the network as well, especially for a multiple layers RNN, the constraint to the range of the recurrent weight with relu activation function can be relaxed to $|u_n|\in [0, \sqrt[\leftroot{-3}\uproot{5}(T-t)]{\gamma}]$. When the recurrent weight is 0, the neuron only uses the information from the current input without keeping any memory from the past. In this way, different neurons can learn to keep memory of different lengths. Note that the regulation on the recurrent weight $u$ is different from the gradient clipping technique. For the gradient clipping or gradient norm clipping \cite{pascanu2013difficulty}, the calculated gradient is already exploded and is forced back to a predefined range. The gradients for the following steps may keep exploding. In this case, the gradient of the other layers relying on this neuron may not be accurate. On the contrary, the regulation proposed here essentially maintains the gradient in an appropriate range without affecting the gradient backprogated through this neuron.  
\vspace{0.1cm}

\begin{figure}[tbp]
	\centering
	\subfigure[]{
	\includegraphics[width=0.8\hsize]{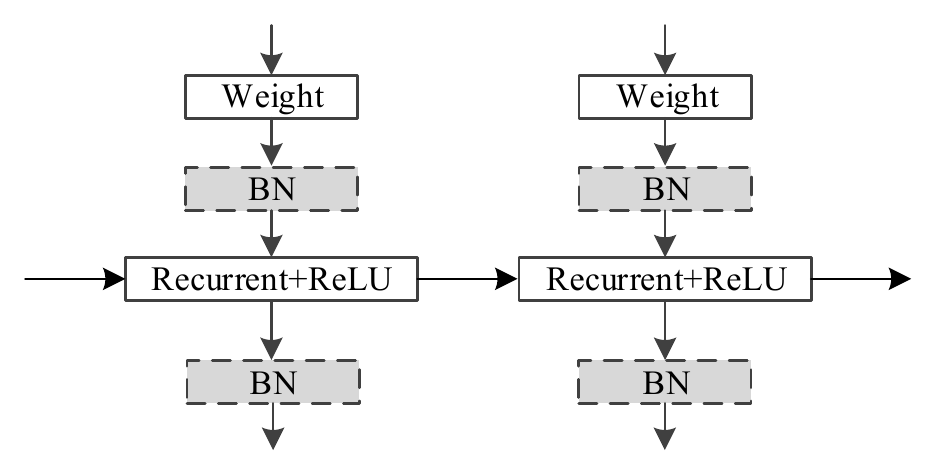}\label{basic}}
	\subfigure[]{
	\includegraphics[width=0.8\hsize]{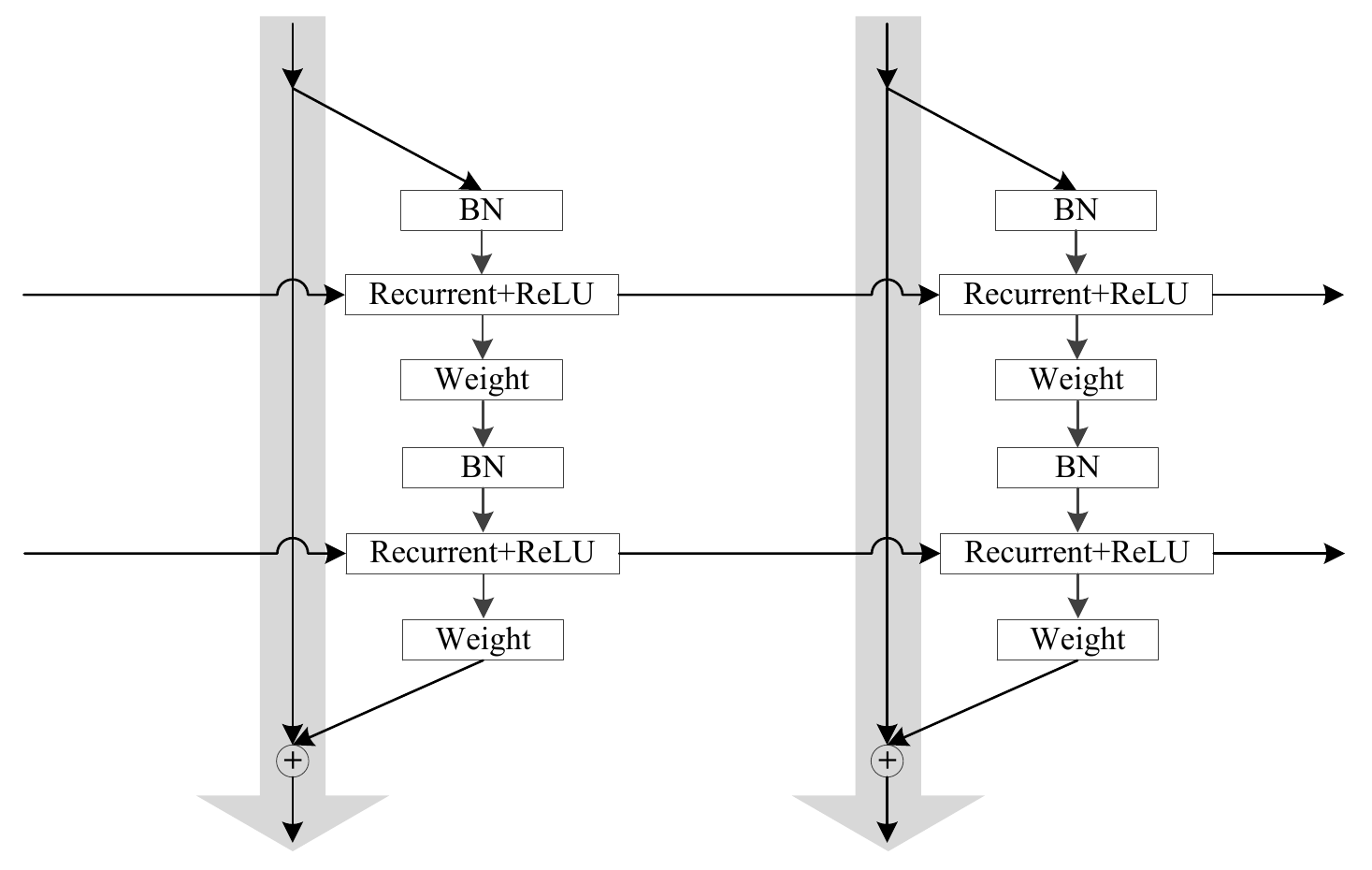}\label{basic_rnnpreres}}
	\caption{Illustration of (a) the basic IndRNN architecture and (b) the residual IndRNN architecture.} 
\label{rnnillustration}
\end{figure}

\section{Multiple-layer IndRNN}
\label{relationRNN}
As mentioned above, neurons in the same IndRNN layer are independent of each other, and cross channel information over time is explored through multiple layers of IndRNNs. To illustrate this, we compare a two-layer IndRNN with a traditional single layer RNN. For simplicity, the bias term is ignored for both IndRNN and traditional RNN. Assume a simple $N$-neuron two-layer network where the recurrent weights for the second layer are zero which means the second layer is just a fully connected layer shared over time. The Hadamard product ($\mathbf{u}\odot\mathbf{h}_{t-1}$) can be represented in the form of matrix product by $diag(u_{1}, u_{2}, \hdots, u_{N})\mathbf{h}_{t-1}$. In the following, $diag(u_{1}, u_{2}, \hdots, u_{N})$ is shortened as $diag(u_{i})$. Assume that the activation function is a linear function $\sigma (x)=x$. The first and second layers of a two-layer IndRNN can be represented by (\ref{firstlayerKIRNN}) and (\ref{secondlayerKIRNN}), respectively. 
\begin{align}
\label{firstlayerKIRNN}
\mathbf{h}_{f,t}&=\mathbf{W}_f\mathbf{x}_{f,t}+diag(u_{fi})\mathbf{h}_{f,t-1}
\\
\label{secondlayerKIRNN}
\mathbf{h}_{s,t}&=\mathbf{W}_s\mathbf{h}_{f,t}
\end{align}
Assuming $\mathbf{W}_s$ is invertible, then 
\begin{equation}
\mathbf{W}_s^{-1}\mathbf{h}_{s,t}=\mathbf{W}_f\mathbf{x}_{f,t}+diag(u_{fi})\mathbf{W}_s^{-1}\mathbf{h}_{s,t-1}
\end{equation}
Thus
\begin{equation}
\mathbf{h}_{s,t}=\mathbf{W}_s\mathbf{W}_f\mathbf{x}_{f,t}+\mathbf{W}_sdiag(u_{fi})\mathbf{W}_s^{-1}\mathbf{h}_{s,t-1}
\label{kirnnreprnn}
\end{equation}

By assigning $\mathbf{U}=\mathbf{W}_sdiag(u_{fi})\mathbf{W}_s^{-1}$ and $\mathbf{W}=\mathbf{W}_s\mathbf{W}_f$, it becomes
\begin{equation}
\mathbf{h}_t=\mathbf{Wx}_t+\mathbf{Uh}_{t-1}
\label{nobrnn}
\end{equation}
which is a traditional RNN. Note that this only imposes the constraint that the recurrent weight ($\mathbf{U}$) is diagonalizable. Therefore, the simple two-layer IndRNN network can represent a traditional RNN network with a diagonalizable recurrent weight ($\mathbf{U}$). In other words, under linear activation, a traditional RNN with a diagonalizable recurrent weight ($\mathbf{U}$) is a special case of a two-layer IndRNN where the recurrent weight of the second layer is zero and the input weight of the second layer is invertible. 

It is known that a non-diagonalizable matrix can be made diagonalizable with a perturbation matrix composed of small entries. A stable RNN network needs to be robust to small perturbations (in order to deal with precision errors for example). It is possible to find an RNN network with a diagonalizable recurrent weight matrix to approximate a stable RNN network with a non-diagonalizable recurrent weight matrix. Therefore, a traditional RNN with a linear activation is a special case of a two-layer IndRNN. For a traditional RNN with a nonlinear activation function, its relationship with the proposed IndRNN is yet to be established theoretically. However, we have shown empirically that the proposed IndRNN can achieve better performance than a traditional RNN with a nonlinear activation function.

Regarding the number of parameters, for a $N$-neuron RNN network with input of dimension $M$, the number of parameters in a traditional RNN is $M\times N+N\times N$, while the number of parameters using one-layer IndRNN is $M\times N+N$. For a two-layer IndRNN where both layers consist of $N$ neurons, the number of parameters is $M\times N+N\times N+2\times N$, which is of a similar order to the traditional RNN. 

In all, the cross-channel information can be well explored with a multiple-layer IndRNN although IndRNN neurons are independent of each other in each layer. 

\subsection{Deeper and Longer IndRNN Architectures}
\label{rnn_arcs}
In the proposed IndRNN, the processing of the input ($\mathbf{Wx}_t+\mathbf{b}$) is independent at different timesteps and can be  implemented in parallel as in \cite{bradbury2016quasi,lei2017training}. The proposed IndRNN can be extended to a convolutional IndRNN where, instead of processing input of each time step using a fully connected weight ($\mathbf{Wx}_t$), it is processed with convolutional operation ($\mathbf{W*x}_t$, where $*$ denotes the convolution operator).

The basic IndRNN architecture is shown in Fig. \ref{basic}, where ``weight'' and ``Recurrent+ReLU'' denote the processing of input and the recurrent process at each step with relu as the activation function. By stacking this basic architecture, a deep IndRNN network can be constructed. Compared with an LSTM-based architecture using the sigmoid and hyperbolic tangent functions decaying the gradient over layers, a non-saturated activation function such as relu reduces the gradient vanishing problem over layers. In addition, batch normalization, denoted as ``BN'', can also be employed in the IndRNN network before or after the activation function as shown in Fig. \ref{basic}.  

Since the weight layer ($\mathbf{Wx}_t+\mathbf{b}$) is used to process the input, it is natural to extend it to multiple layers to deepen the processing. Also the layers used to process the input can be of the residual structures in the same way as in CNN \cite{he2016deep}. With the simple structure of IndRNN, it is very easy to extend it to different networks architectures. For example, in addition to simply stacking IndRNNs or stacking the layers for processing the input, IndRNNs can also be stacked in the form of residual connections. Fig. \ref{basic_rnnpreres} shows an example of a residual IndRNN based on the ``pre-activation'' type of residual layers in \cite{he2016identity}. At each time step, the gradient can be directly propagated to the other layers from the identity mapping. Since IndRNN addresses the gradient exploding and vanishing problems over time, the gradient can be efficiently propagated over different time steps. Therefore, the network can be substantially deeper and longer. The deeper and longer IndRNN network can be trained end-to-end similarly as other networks.

\begin{figure*}[tbp]
	\centering
	\subfigure[]{
	\includegraphics[width=0.43\hsize]{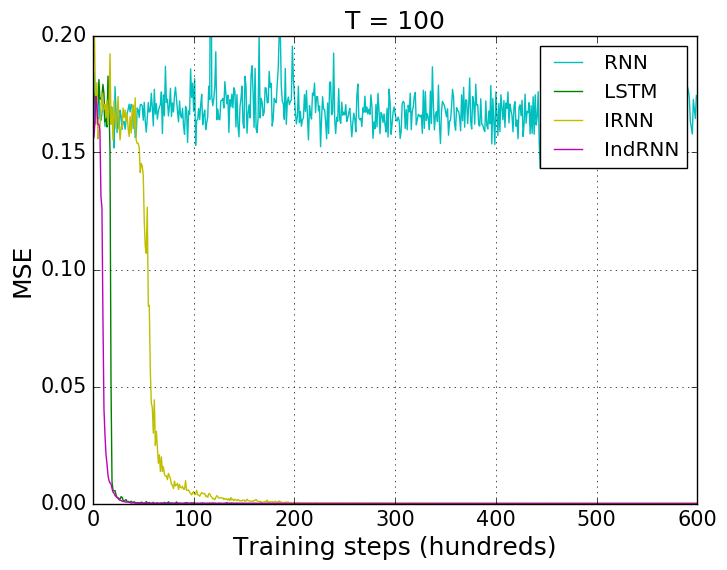}\label{adding_100}}
	\hspace{0.5cm}
	\subfigure[]{
	\includegraphics[width=0.43\hsize]{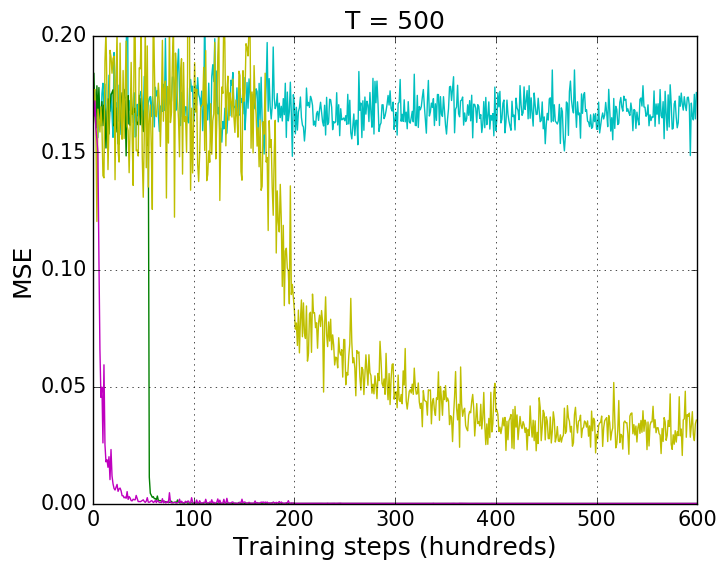}\label{adding_500}}
	\subfigure[]{
	\includegraphics[width=0.43\hsize]{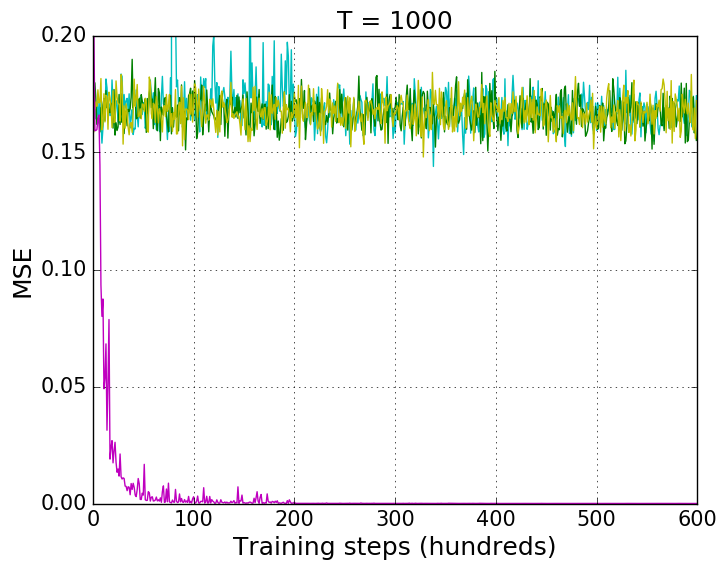}\label{adding_1000}}
	\hspace{0.5cm}
	\subfigure[]{
	\includegraphics[width=0.43\hsize]{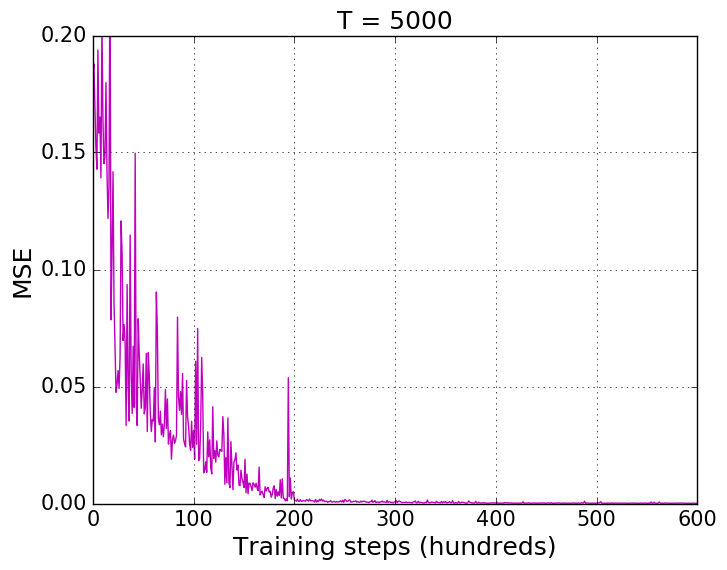}\label{adding_5000}}
	\caption{Results of the adding problem for different sequence lengths. The legends for all figures are the same and thus only shown in (a).} 
	\label{adding}
\end{figure*}

\section{Experiments}
In this Section, evaluation of the proposed IndRNN on various tasks are presented.

\subsection{Adding Problem}
\label{addingdesc}
The adding problem~\cite{hochreiter1997long, arjovsky2015unitary} is commonly used to evaluate the performance of RNN models. Two sequences of length $T$ are taken as input. The first sequence is uniformly sampled in the range $(0,1)$ while the second sequence consists of two entries being $1$ and the rest being $0$. The output is the sum of the two entries in the first sequence indicated by the two entries of $1$ in the second sequence. Three different lengths of sequences, $T=100$, $500$ and $1000$, were used for the experiments to show whether the tested models have the ability to model long-term memory.
 
The RNN models included in the experiments for comparison are the traditional RNN with tanh, LSTM, IRNN (RNN with relu). The proposed IndRNN was evaluated with relu activation function. Since GRU achieved similar performance as LSTM \cite{jozefowicz2015empirical}, it is not included in the report. RNN, LSTM, and IRNN are all one layer while the IndRNN model is two layers. $128$ hidden units were used for all the models, and the number of parameters for RNN, LSTM, and two-layer IndRNN are $16K$, $67K$ and $17K$, respectively. It can be seen that the two-layer IndRNN has a comparable number of parameters to that of the one-layer RNN, while many more parameters are needed for LSTM. As discussed in Subsection \ref{BPTT}, the recurrent weight is constrained in the range of $|u_n|\in (0, \sqrt[\leftroot{-3}\uproot{5}T]{2})$ for the IndRNN. 

Mean squared error (MSE) was used as the objective function and the Adam optimization method \cite{kingma2014adam} was used for training. The baseline performance (predicting 1 as the output regardless of the input sequence) is mean squared error of 0.167 (the variance of the sum of two independent uniform distributions). The initial learning rate was set to $2\times 10^{-3}$ for models with tanh activation and set as $2\times 10^{-4}$ for models with relu activations. However, as the length of the sequence increases, the IRNN model do not converge and thus a smaller initial learning rate ($10^{-5}$) was used. The learning rate was reduced by a factor of 10 every 20K training steps. The training data and testing data were all generated randomly throughout the experiments, different from \cite{arjovsky2015unitary} which only used a set of randomly pre-generated data. 

The results are shown in Fig. \ref{adding_100}, \ref{adding_500} and \ref{adding_1000}. First, for short sequences ($T=100$), most of the models (except RNN with tanh) performed well as they converged to a very small error (much smaller than the baseline). When the length of the sequences increases, the IRNN and LSTM models have difficulties in converging, and when the sequence length reaches $1000$, IRNN and LSTM cannot minimize the error any more. However, the proposed IndRNN can still converge to a small error very quickly. This indicates that the proposed IndRNN can model a longer-term memory than the traditional RNN and LSTM. 

From the figures, it can also be seen that the traditional RNN and LSTM can only keep a mid-range memory (about 500 - 1000 time steps). To evaluate the proposed IndRNN model for very long-term memory, experiments on sequences with length $5000$ were conducted where the result is shown in Fig. \ref{adding_5000}. It can be seen that IndRNN can still model it very well. Note that the noise in the result of IndRNN is because the initial learning rate ($2\times 10^{-4}$) was relatively large and once the learning rate dropped, the performance became robust. This demonstrates that IndRNN can effectively address the gradient exploding and vanishing problem over time and keep a long-term memory.

\subsubsection{Analysis of Neurons' Behaviour}
\label{analysisneuron}
\begin{figure}[tbp]
	\centering
	\subfigure[]{
	\includegraphics[width=0.8\hsize]{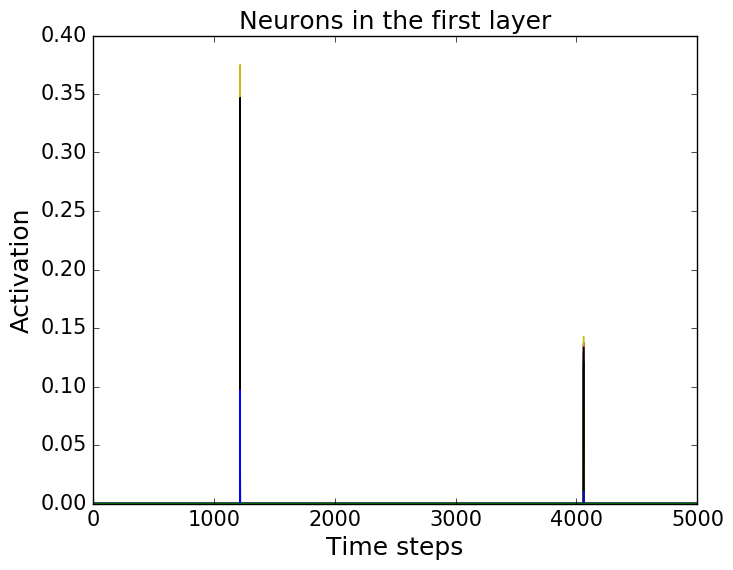}\label{1layer}}
	\hspace{1cm}
	\subfigure[]{
	\includegraphics[width=0.8\hsize]{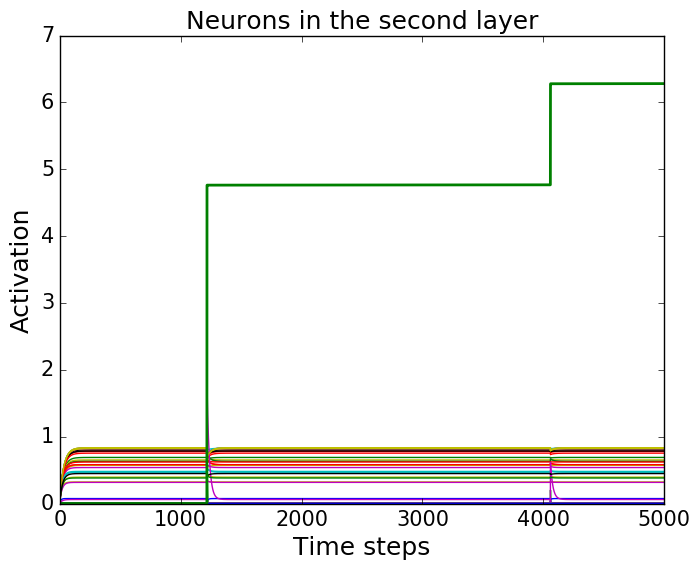}\label{2layer}}
	\caption{Neurons' behaviour in different layers of the proposed IndRNN for long sequences (5000 time steps) in the adding problem.} 
	\label{neuronbehav}
\end{figure}

In the proposed IndRNN, neurons in each layer are independent of each other which allows analysis of each neuron's behaviour without considering the effect coming from other neurons. Fig. \ref{1layer} and \ref{2layer} show the activation of the neurons in the first and second layers, respectively, for one random input with sequence length $5000$. It can be seen that neurons in the first layer mainly pick up the information of the numbers to be added, where the strong responses correspond to the locations to be summed indicated by the sequence. It can be regarded as reducing noise, i.e., reducing the effect of other non-useful inputs in the sequence. For the second layer, one neuron aggregates inputs to long-term memory while others generally preserve their own state or process short-term memory which may not be useful in the testing case (since only the hidden state of the last time step is used as output). From this result, we conjecture that only one neuron is needed in the second layer to model the adding problem. Moreover, since neurons in the second layer are independent from each other, one neuron can still work with the others removed (which is not possible for the traditional RNN models).

\begin{figure}[tbp]
\begin{minipage}[t]{1\linewidth}
\centering
    \includegraphics[width=0.8\linewidth]{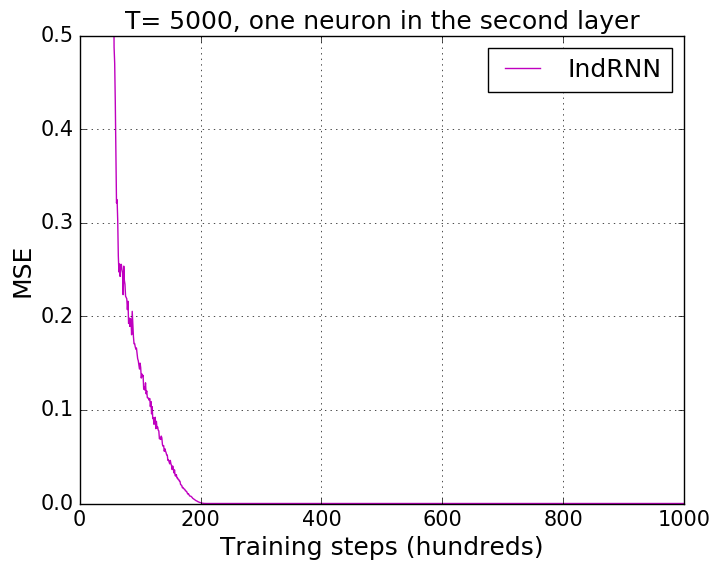}
    \caption{Result of the adding problem with just one neuron in the second layer for sequences of length 5000.} 
    \label{resultoneneuron}
\end{minipage}%
\end{figure}

To verify the above conjecture, an experiment was conducted where the first IndRNN layer is initialized with the trained weights and the second IndRNN layer only consists of one neuron initialized with the weight of the neuron that keeps the long-term memory. Accordingly, the final fully connected layer used for output is a neuron with only one input and one output, i.e., two scalar values including one weight parameter and one bias parameter. Only the final output layer was trained/fine-tuned in this experiment and the result is shown in Fig. \ref{resultoneneuron}. It can be seen that with only one IndRNN neuron in the second layer, the model is still able to model the adding problem very well for sequences with length $5000$ as expected.


\subsection{Sequential MNIST Classification}
Sequential MNIST classification is another problem that is widely used to evaluate RNN models. The pixels of MNIST digits \cite{lecun1998gradient} are presented sequentially to the networks and classification is performed after reading all pixels. To make the task even harder, the permuted MNIST classification was also used where the pixels are processed with a fixed random permutation. Since an RNN with tanh does not converge to a high accuracy (as reported in the literature \cite{le2015simple}), only IndRNN with relu was evaluated. As explained in Section \ref{rnn_arcs}, IndRNN can be stacked into a deep network. Here we used a six-layer IndRNN, and each layer has 128 neurons. To accelerate the training, batch normalization is inserted after each layer. The Adam optimization was used with the initial learning rate $2\times 10^{-4}$ and reduced by a factor of 10 every 600K training steps. The results are shown in Table \ref{result_mnist} in comparison with the existing methods. It can be seen that IndRNN achieved better performance than the existing RNN models. 

\begin{table}
\centering
\caption{Results (in terms of error rate (\%)) for the sequential MNIST and permuted MNIST.} 
  \begin{tabular}{@{}l@{}cc}
  \hline
   & MNIST & pMNIST \\
  \hline
  IRNN \cite{le2015simple} & $5.0$ & $18$ \\
  uRNN \cite{arjovsky2015unitary} & $4.9$ & $8.6$\\
  RNN-path \cite{neyshabur2016path} & $3.1$ & - \\
  LSTM \cite{arjovsky2015unitary} & $1.8$ & $12$ \\ 
  LSTM+Recurrent dropout \cite{semeniuta2016recurrent} & - & $7.5$ \\ 
  LSTM+Recurrent batchnorm \cite{cooijmans2016recurrent} & - & $4.6$ \\ 
  LSTM+Zoneout \cite{krueger2016zoneout} & - & $6.9$ \\ 
  LSTM+Recurrent batchnorm+Zoneout & - & $4.1$ \\ 
  \hline  
  \textbf{IndRNN (6 layers)} & $1.0$ & $4.0$\\
  \hline
  \end{tabular}
\label{result_mnist}
\end{table}

\subsection{Language Modeling}
\subsubsection{Char-level Penn Treebank}
In this subsection, we evaluate the performance of the proposed IndRNN on the language modelling task using the character-level Penn Treebank (PTB-c) dataset. The test setting is similar to \cite{cooijmans2016recurrent}. A six-layer IndRNN with $2000$ hidden neurons is used for the test. To demonstrate that the IndRNN network can be very deep with the residual connections, a 21-layer residual IndRNN as shown in Fig. \ref{basic_rnnpreres} in Subsection \ref{rnn_arcs} was adopted. The frame-wise batch normalization \cite{laurent2016batch} is applied, and the batch size is set to $128$. Adam was used for training with initial learning rate set to $2 \times 10^{-4}$ and dropped by a factor of 5 when performance on the validation set was no longer improved (with patience 20). Dropout \cite{gal2016theoretically} with a dropping probability of 0.25 and 0.3 were used for the 6-layer IndRNN and the residual IndRNN. The sequences are non-overlapping and length $T=50$ and $T=150$ were both tested in training and testing.

The results are shown in Table \ref{result_penntree} in comparison with the existing methods. Performance was evaluated using bits per character metric (BPC). It can be seen that the proposed IndRNN model achieved better performance than the traditional RNN and LSTM models. It can also been seen that with a deeper residual IndRNN, the performance can be further improved. Also an improvement can be achieved with longer temporal dependencies (from time step 50 to 150) as shown in Table \ref{result_penntree}. 

\begin{table}
\centering
\tabcolsep=12pt
\caption{Results of char-level PTB for our proposed IndRNN model in comparison with results reported in the literature, in terms of BPC.}\label{result_penntree}
\begin{threeparttable}
  \begin{tabular}{l c}
  \hline
   & Test \\
  \hline
  RNN-tanh \cite{krueger2016regularizing} & $1.55$ \\
  RNN-relu \cite{neyshabur2016path} &  $1.55$ \\
  RNN-TRec \cite{krueger2016regularizing} & $1.48$ \\
  RNN-path \cite{neyshabur2016path} & $1.47$ \\
  HF-MRNN \cite{mikolov2012subword} & $1.42$ \\ 
  LSTM \cite{krueger2016zoneout}  & $1.36$ \\
  LSTM+Recurrent dropout \cite{semeniuta2016recurrent} & $1.32$ \\  
  LSTM+Recurrent batchnorm \cite{cooijmans2016recurrent} & $1.32$ \\
  LSTM+Zoneout \cite{krueger2016zoneout} &  $1.27$ \\
  HyperLSTM + LN \cite{ha2016hypernetworks} & $1.25$ \\
  Hierarchical Multiscale LSTM + LN \cite{chung2016hierarchical} & $1.24$ \\
  Fast-slow LSTM \cite{mujika2017fast} & $1.19$ \\
  Neural Architecture Search \cite{zoph2016neural} & $1.21$ \\
  \hline  
  \textbf{IndRNN (6 layers, 50 steps)} & $1.26$ \\
  \hline 
  \textbf{IndRNN (6 layers, 150 steps)} & $1.23$ \\
  \hline 
  \textbf{res-IndRNN (21 layers, 50 steps)} & $1.21$ \\
  \hline  
  \textbf{res-IndRNN (11 layers\textsuperscript{*}, 150 steps)} & $1.19$ \\
  \hline
  \end{tabular}
      \begin{tablenotes}
      \small
      \item \textsuperscript{*}Note that due to the limitation of GPU memory, an 11-layer residual IndRNN was used for time step 150 instead of 21 layers.
    \end{tablenotes}
\label{result_final}
\end{threeparttable}
\end{table}

\subsubsection{Word-level Penn Treebank}
In this subsection, the performance of the proposed IndRNN on the word-level Penn Treebank dataset is evaluated. The test setting is similar to \cite{krueger2016zoneout}. A 11-layer residual IndRNN was used for test and the weight tying \cite{inan2016tying,press2016using} of the input embedding and the final output weight is also adopted. The frame-wise batch normalization \cite{laurent2016batch} is applied, and the batch size is set to $128$. Adam was used for training with initial learning rate set to $5 \times 10^{-4}$ and dropped by a factor of 5 when performance on the validation set was no longer improved (with patience 20). The sequences are non-overlapping and length $T=50$ was used in training and testing. Dropout \cite{gal2016theoretically} with a dropping probability of 0.35 were used among IndRNN layers (including embedding) while 0.8 is used after the last IndRNN layer. The recurrent weights are initialized with $Normal(0.4, 0.2)$, which makes the network starts with learning more mid-range memory.

The results are shown in Table \ref{result_wPTB} in comparison with the existing methods. It can be seen that the proposed IndRNN model achieved better performance than most of the traditional RNN and LSTM models except the neural architecture search \cite{zoph2016neural} which constructs new models while learning. 

\begin{table}
\centering
\tabcolsep=12pt
\caption{Results of word-level PTB for our proposed IndRNN model in comparison with results reported in the literature, in terms of perplexity.}\label{result_wPTB}
  \begin{tabular}{l c}
  \hline
   & Test \\
  \hline
  RNN-LDA + KN-5 + cache \cite{mikolov2012context}  & $92.0$ \\
  Deep RNN \cite{pascanu2013construct}  & $107.5$ \\
  CharCNN \cite{kim2016character}  & $78.9$ \\
  LSTM \cite{krueger2016zoneout}  & $114.5$ \\
  LSTM+Recurrent dropout \cite{semeniuta2016recurrent} & $87.0$ \\  
  LSTM+Zoneout \cite{krueger2016zoneout} &  $77.4$ \\
  LSTM+Variational Dropout \cite{gal2016theoretically} & $73.4$ \\
  Pointer Sentinel LSTM \cite{merity2016pointer} & $70.9$ \\
  RHN \cite{zilly2016recurrent} & $65.4$ \\
  Neural Architecture Search \cite{zoph2016neural} & $62.4$ \\
  \hline  
  \textbf{res-IndRNN (11 layers)} & $65.3$ \\
  \hline
  \end{tabular}
\end{table}

\subsection{Skeleton based Action Recognition}
The NTU RGB+D dataset \cite{shahroudy2016ntu} was used for the skeleton based action recognition. This dataset is currently the largest action recognition dataset with skeleton modality. It contains $56880$ sequences of 60 action classes, including Cross-Subject (CS) (40320 and 16560 samples for training and testing, respectively) and Cross-View (CV) (37920 and 18960 samples for training and testing, respectively) evaluation protocols \cite{shahroudy2016ntu}. In each evaluation protocol, 5\% of the training data was used for evaluation as suggested in \cite{shahroudy2016ntu} and 20 frames were sampled from each instance as one input in the same way as in \cite{liu2016spatio}. The joint coordinates of two subject skeletons were used as input. If only one is present, the second was set as zero. For this dataset, when multiple skeletons are present in the scene, the skeleton identity captured by the Kinect sensor may be changed over time. Therefore, an alignment process was first applied to keep the same skeleton saved in the same data array over time. A four-layer IndRNN and a six-layer IndRNN with $512$ hidden neurons were both tested. Batch size was 128 and the Adam optimization was used with the initial learning rate $2\times 10^{-4}$ and decayed by $10$ once the evaluation accuracy does not increase. Dropout \cite{gal2016theoretically} was applied after each IndRNN layer with a dropping probability of $0.25$ and $0.1$ for CS and CV settings, respectively. 

The final result is shown in Table \ref{result_ntu} including comparisons with the existing methods. It can be seen that the proposed IndRNN greatly improves the performance over other RNN or LSTM models on the same task. For CS, RNN and LSTM of 2 layers can only achieve accuracies of 56.29\% and 60.09\% while a 4-layer IndRNN achieved 78.58\%. For CV, RNN and LSTM of 2 layers only achieved accuracies of 64.09\% and 67.29\% while 4-layer IndRNN achieved 83.75\%. As demonstrated in \cite{liu2016spatio,shahroudy2016ntu}, the performance of LSTM cannot be further improved by simply increasing the number of parameters or increasing the number of layers. However, by increasing the 4-layer IndRNN to a 6-layer IndRNN, the performance is further improved to 81.80\% and 87.97\% for CS and CV, respectively. This performance is better than the state-of-the-art methods including those with attention models \cite{song2017end,baradel2017pose} and other techniques \cite{zhang2017geometric,liu2016spatio}.

\begin{table}
\caption{Results of all skeleton based methods on NTU RGB+D dataset.}
\label{result_ntu}
\begin{center}
  \begin{tabular}{l c c}
  \hline
  Method & CS & CV \\
  \hline
  Deep learning on Lie Group \cite{huang2016deep} & 61.37\% & 66.95\%  \\
  \hline
  JTM+CNN \cite{wang2016action} & 73.40\% & 75.20\%  \\
  \hline
  Res-TCN \cite{kim2017interpretable} & 74.30\% & 83.10\%  \\
  \hline
  SkeletonNet(CNN) \cite{ke2017skeletonnet} & 75.94\% & 81.16\%  \\
  \hline
  JDM+CNN \cite{li2017joint} & 76.20\% & 82.30\%  \\
  \hline
  Clips+CNN+MTLN \cite{ke2017new} & 79.57\% & 84.83\%  \\
  \hline
  Enhanced Visualization+CNN \cite{liu2017enhanced} & 80.03\% & 87.21\%  \\
  \hline
  1 Layer RNN \cite{shahroudy2016ntu} & 56.02\% & 60.24\%  \\
  \hline
  2 Layer RNN \cite{shahroudy2016ntu} & 56.29\% & 64.09\%  \\
  \hline
  1 Layer LSTM \cite{shahroudy2016ntu} & 59.14\% & 66.81\%  \\
  \hline
  2 Layer LSTM \cite{shahroudy2016ntu} & 60.09\% & 67.29\%  \\
  \hline
  1 Layer PLSTM \cite{shahroudy2016ntu} & 62.05\% & 69.40\%  \\
  \hline
  2 Layer PLSTM \cite{shahroudy2016ntu} & 62.93\% & 70.27\%  \\
  \hline
  JL\_d+RNN \cite{zhang2017geometric} & 70.26\% & 82.39\%  \\
  \hline
  STA-LSTM \cite{song2017end} & 73.40\% & 81.20\%  \\
  \hline
  ST-LSTM + Trust Gate \cite{liu2016spatio} & 69.20\% & 77.70\%  \\
  \hline
  Pose conditioned STA-LSTM\cite{baradel2017pose} & 77.10\% & 84.50\%  \\
  \hline
  \textbf{IndRNN (4 layers)} & 78.58\% & 83.75\%  \\
  \hline
  \textbf{IndRNN (6 layers)} & 81.80\% & 87.97\%  \\
  \hline
  \end{tabular}
\end{center}
\end{table}

\section{Conclusion}
In this paper, we presented an independently recurrent neural network (IndRNN), where neurons in one layer are independent of each other. The gradient backpropagation through time process for the IndRNN has been explained and a regulation technique has been developed to effectively address the gradient vanishing and exploding problems. Compared with the existing RNN models including LSTM and GRU, IndRNN can process much longer sequences. The basic IndRNN can be stacked to construct a deep network especially combined with residual connections over layers, and the deep network can be trained robustly. In addition, independence among neurons in each layer allows better interpretation of the neurons. Experiments on multiple fundamental tasks have verified the advantages of the proposed IndRNN over existing RNN models. 

{\small
\bibliographystyle{ieee}
\bibliography{reRNNreference}

\begin{thebibliography}{10}\itemsep=-1pt

\bibitem{arjovsky2015unitary}
M.~Arjovsky, A.~Shah, and Y.~Bengio.
\newblock Unitary evolution recurrent neural networks.
\newblock {\em arXiv preprint arXiv:1511.06464}, 2015.

\bibitem{baradel2017pose}
F.~Baradel, C.~Wolf, and J.~Mille.
\newblock Pose-conditioned spatio-temporal attention for human action
  recognition.
\newblock {\em arXiv preprint arXiv:1703.10106}, 2017.

\bibitem{bradbury2016quasi}
J.~Bradbury, S.~Merity, C.~Xiong, and R.~Socher.
\newblock Quasi-recurrent neural networks.
\newblock {\em arXiv preprint arXiv:1611.01576}, 2016.

\bibitem{byeon2015scene}
W.~Byeon, T.~M. Breuel, F.~Raue, and M.~Liwicki.
\newblock Scene labeling with lstm recurrent neural networks.
\newblock In {\em Proceedings of the IEEE Conference on Computer Vision and
  Pattern Recognition}, pages 3547--3555, 2015.

\bibitem{cho2014learning}
K.~Cho, B.~Van~Merri{\"e}nboer, C.~Gulcehre, D.~Bahdanau, F.~Bougares,
  H.~Schwenk, and Y.~Bengio.
\newblock Learning phrase representations using rnn encoder-decoder for
  statistical machine translation.
\newblock {\em arXiv preprint arXiv:1406.1078}, 2014.

\bibitem{chung2016hierarchical}
J.~Chung, S.~Ahn, and Y.~Bengio.
\newblock Hierarchical multiscale recurrent neural networks.
\newblock {\em arXiv preprint arXiv:1609.01704}, 2016.

\bibitem{cooijmans2016recurrent}
T.~Cooijmans, N.~Ballas, C.~Laurent, {\c{C}}.~G{\"u}l{\c{c}}ehre, and
  A.~Courville.
\newblock Recurrent batch normalization.
\newblock {\em arXiv preprint arXiv:1603.09025}, 2016.

\bibitem{donahue2015long}
J.~Donahue, L.~Anne~Hendricks, S.~Guadarrama, M.~Rohrbach, S.~Venugopalan,
  K.~Saenko, and T.~Darrell.
\newblock Long-term recurrent convolutional networks for visual recognition and
  description.
\newblock In {\em Proceedings of the IEEE Conference on Computer Vision and
  Pattern Recognition}, pages 2625--2634, 2015.

\bibitem{gal2016theoretically}
Y.~Gal and Z.~Ghahramani.
\newblock A theoretically grounded application of dropout in recurrent neural
  networks.
\newblock In {\em Advances in neural information processing systems}, pages
  1019--1027, 2016.

\bibitem{greff2017lstm}
K.~Greff, R.~K. Srivastava, J.~Koutn{\'\i}k, B.~R. Steunebrink, and
  J.~Schmidhuber.
\newblock Lstm: A search space odyssey.
\newblock {\em IEEE transactions on neural networks and learning systems},
  2017.

\bibitem{ha2016hypernetworks}
D.~Ha, A.~Dai, and Q.~V. Le.
\newblock Hypernetworks.
\newblock {\em arXiv preprint arXiv:1609.09106}, 2016.

\bibitem{he2016deep}
K.~He, X.~Zhang, S.~Ren, and J.~Sun.
\newblock Deep residual learning for image recognition.
\newblock In {\em Proceedings of the IEEE conference on computer vision and
  pattern recognition}, pages 770--778, 2016.

\bibitem{he2016identity}
K.~He, X.~Zhang, S.~Ren, and J.~Sun.
\newblock Identity mappings in deep residual networks.
\newblock In {\em European Conference on Computer Vision}, pages 630--645.
  Springer, 2016.

\bibitem{hochreiter1997long}
S.~Hochreiter and J.~Schmidhuber.
\newblock Long short-term memory.
\newblock {\em Neural computation}, 9(8):1735--1780, 1997.

\bibitem{huang2016deep}
Z.~Huang, C.~Wan, T.~Probst, and L.~Van~Gool.
\newblock Deep learning on lie groups for skeleton-based action recognition.
\newblock {\em arXiv preprint arXiv:1612.05877}, 2016.

\bibitem{inan2016tying}
H.~Inan, K.~Khosravi, and R.~Socher.
\newblock Tying word vectors and word classifiers: A loss framework for
  language modeling.
\newblock {\em arXiv preprint arXiv:1611.01462}, 2016.

\bibitem{jordan1997serial}
M.~I. Jordan.
\newblock Serial order: A parallel distributed processing approach.
\newblock {\em Advances in psychology}, 121:471--495, 1997.

\bibitem{jozefowicz2015empirical}
R.~Jozefowicz, W.~Zaremba, and I.~Sutskever.
\newblock An empirical exploration of recurrent network architectures.
\newblock In {\em Proceedings of the 32nd International Conference on Machine
  Learning (ICML-15)}, pages 2342--2350, 2015.

\bibitem{karpathy2015visualizing}
A.~Karpathy, J.~Johnson, and L.~Fei-Fei.
\newblock Visualizing and understanding recurrent networks.
\newblock {\em arXiv preprint arXiv:1506.02078}, 2015.

\bibitem{ke2017skeletonnet}
Q.~Ke, S.~An, M.~Bennamoun, F.~Sohel, and F.~Boussaid.
\newblock Skeletonnet: Mining deep part features for 3-d action recognition.
\newblock {\em IEEE Signal Processing Letters}, 24(6):731--735, 2017.

\bibitem{ke2017new}
Q.~Ke, M.~Bennamoun, S.~An, F.~Sohel, and F.~Boussaid.
\newblock A new representation of skeleton sequences for 3d action recognition.
\newblock {\em arXiv preprint arXiv:1703.03492}, 2017.

\bibitem{kim2017interpretable}
T.~S. Kim and A.~Reiter.
\newblock Interpretable 3d human action analysis with temporal convolutional
  networks.
\newblock {\em arXiv preprint arXiv:1704.04516}, 2017.

\bibitem{kim2016character}
Y.~Kim, Y.~Jernite, D.~Sontag, and A.~M. Rush.
\newblock Character-aware neural language models.
\newblock In {\em AAAI}, pages 2741--2749, 2016.

\bibitem{kingma2014adam}
D.~P. Kingma and J.~Ba.
\newblock Adam: A method for stochastic optimization.
\newblock {\em arXiv preprint arXiv:1412.6980}, 2014.

\bibitem{krueger2016zoneout}
D.~Krueger, T.~Maharaj, J.~Kram{\'a}r, M.~Pezeshki, N.~Ballas, N.~R. Ke,
  A.~Goyal, Y.~Bengio, H.~Larochelle, A.~Courville, et~al.
\newblock Zoneout: Regularizing rnns by randomly preserving hidden activations.
\newblock {\em arXiv preprint arXiv:1606.01305}, 2016.

\bibitem{krueger2016regularizing}
D.~Krueger and R.~Memisevic.
\newblock Regularizing rnns by stabilizing activations.
\newblock In {\em Proceeding of the International Conference on Learning
  Representations}, 2016.

\bibitem{laurent2016batch}
C.~Laurent, G.~Pereyra, P.~Brakel, Y.~Zhang, and Y.~Bengio.
\newblock Batch normalized recurrent neural networks.
\newblock In {\em Acoustics, Speech and Signal Processing (ICASSP), 2016 IEEE
  International Conference on}, pages 2657--2661. IEEE, 2016.

\bibitem{le2015simple}
Q.~V. Le, N.~Jaitly, and G.~E. Hinton.
\newblock A simple way to initialize recurrent networks of rectified linear
  units.
\newblock {\em arXiv preprint arXiv:1504.00941}, 2015.

\bibitem{lecun1998gradient}
Y.~LeCun, L.~Bottou, Y.~Bengio, and P.~Haffner.
\newblock Gradient-based learning applied to document recognition.
\newblock {\em Proceedings of the IEEE}, 86(11):2278--2324, 1998.

\bibitem{lei2017training}
T.~Lei and Y.~Zhang.
\newblock Training rnns as fast as cnns.
\newblock {\em arXiv preprint arXiv:1709.02755}, 2017.

\bibitem{li2017joint}
C.~Li, Y.~Hou, P.~Wang, and W.~Li.
\newblock Joint distance maps based action recognition with convolutional
  neural networks.
\newblock {\em IEEE Signal Processing Letters}, 24(5):624--628, 2017.

\bibitem{liu2016spatio}
J.~Liu, A.~Shahroudy, D.~Xu, and G.~Wang.
\newblock Spatio-temporal lstm with trust gates for 3d human action
  recognition.
\newblock In {\em European Conference on Computer Vision}, pages 816--833.
  Springer, 2016.

\bibitem{liu2017enhanced}
M.~Liu, H.~Liu, and C.~Chen.
\newblock Enhanced skeleton visualization for view invariant human action
  recognition.
\newblock {\em Pattern Recognition}, 68:346--362, 2017.

\bibitem{merity2016pointer}
S.~Merity, C.~Xiong, J.~Bradbury, and R.~Socher.
\newblock Pointer sentinel mixture models.
\newblock {\em arXiv preprint arXiv:1609.07843}, 2016.

\bibitem{mikolov2012subword}
T.~Mikolov, I.~Sutskever, A.~Deoras, H.-S. Le, S.~Kombrink, and J.~Cernocky.
\newblock Subword language modeling with neural networks.
\newblock {\em preprint (http://www. fit. vutbr. cz/imikolov/rnnlm/char. pdf)},
  2012.

\bibitem{mikolov2012context}
T.~Mikolov and G.~Zweig.
\newblock Context dependent recurrent neural network language model.
\newblock {\em SLT}, 12:234--239, 2012.

\bibitem{mujika2017fast}
A.~Mujika, F.~Meier, and A.~Steger.
\newblock Fast-slow recurrent neural networks.
\newblock In {\em Advances in Neural Information Processing Systems}, pages
  5917--5926, 2017.

\bibitem{neyshabur2016path}
B.~Neyshabur, Y.~Wu, R.~R. Salakhutdinov, and N.~Srebro.
\newblock Path-normalized optimization of recurrent neural networks with relu
  activations.
\newblock In {\em Advances in Neural Information Processing Systems}, pages
  3477--3485, 2016.

\bibitem{parlett1964laguerre}
B.~Parlett.
\newblock Laguerre's method applied to the matrix eigenvalue problem.
\newblock {\em Mathematics of Computation}, 18(87):464--485, 1964.

\bibitem{pascanu2013construct}
R.~Pascanu, C.~Gulcehre, K.~Cho, and Y.~Bengio.
\newblock How to construct deep recurrent neural networks.
\newblock {\em arXiv preprint arXiv:1312.6026}, 2013.

\bibitem{pascanu2013difficulty}
R.~Pascanu, T.~Mikolov, and Y.~Bengio.
\newblock On the difficulty of training recurrent neural networks.
\newblock In {\em International Conference on Machine Learning}, pages
  1310--1318, 2013.

\bibitem{pradhanexploring}
S.~Pradhan and S.~Longpre.
\newblock Exploring the depths of recurrent neural networks with stochastic
  residual learning.
\newblock {\em Report}.

\bibitem{press2016using}
O.~Press and L.~Wolf.
\newblock Using the output embedding to improve language models.
\newblock In {\em Proceedings of the 15th Conference of the European Chapter of
  the Association for Computational Linguistics}, volume~2, pages 157--163,
  2017.

\bibitem{qiu2017learning}
Z.~Qiu, T.~Yao, and T.~Mei.
\newblock Learning spatio-temporal representation with pseudo-3d residual
  networks.
\newblock In {\em Proceedings of the IEEE Conference on Computer Vision and
  Pattern Recognition}, pages 5533--5541, 2017.

\bibitem{semeniuta2016recurrent}
S.~Semeniuta, A.~Severyn, and E.~Barth.
\newblock Recurrent dropout without memory loss.
\newblock {\em arXiv preprint arXiv:1603.05118}, 2016.

\bibitem{shahroudy2016ntu}
A.~Shahroudy, J.~Liu, T.-T. Ng, and G.~Wang.
\newblock Ntu rgb+ d: A large scale dataset for 3d human activity analysis.
\newblock In {\em Proceedings of the IEEE Conference on Computer Vision and
  Pattern Recognition}, pages 1010--1019, 2016.

\bibitem{song2017end}
S.~Song, C.~Lan, J.~Xing, W.~Zeng, and J.~Liu.
\newblock An end-to-end spatio-temporal attention model for human action
  recognition from skeleton data.
\newblock In {\em AAAI}, pages 4263--4270, 2017.

\bibitem{talathi2015improving}
S.~S. Talathi and A.~Vartak.
\newblock Improving performance of recurrent neural network with relu
  nonlinearity.
\newblock {\em arXiv preprint arXiv:1511.03771}, 2015.

\bibitem{wang2016action}
P.~Wang, Z.~Li, Y.~Hou, and W.~Li.
\newblock Action recognition based on joint trajectory maps using convolutional
  neural networks.
\newblock In {\em Proceedings of the 2016 ACM on Multimedia Conference}, pages
  102--106. ACM, 2016.

\bibitem{wisdom2016full}
S.~Wisdom, T.~Powers, J.~Hershey, J.~Le~Roux, and L.~Atlas.
\newblock Full-capacity unitary recurrent neural networks.
\newblock In {\em Advances in Neural Information Processing Systems}, pages
  4880--4888, 2016.

\bibitem{wu2016google}
Y.~Wu, M.~Schuster, Z.~Chen, Q.~V. Le, M.~Norouzi, W.~Macherey, M.~Krikun,
  Y.~Cao, Q.~Gao, K.~Macherey, et~al.
\newblock Google's neural machine translation system: Bridging the gap between
  human and machine translation.
\newblock {\em arXiv preprint arXiv:1609.08144}, 2016.

\bibitem{zhang2017geometric}
S.~Zhang, X.~Liu, and J.~Xiao.
\newblock On geometric features for skeleton-based action recognition using
  multilayer lstm networks.
\newblock In {\em Applications of Computer Vision (WACV), 2017 IEEE Winter
  Conference on}, pages 148--157. IEEE, 2017.

\bibitem{zilly2016recurrent}
J.~G. Zilly, R.~K. Srivastava, J.~Koutn{\'\i}k, and J.~Schmidhuber.
\newblock Recurrent highway networks.
\newblock {\em arXiv preprint arXiv:1607.03474}, 2016.

\bibitem{zoph2016neural}
B.~Zoph and Q.~V. Le.
\newblock Neural architecture search with reinforcement learning.
\newblock {\em arXiv preprint arXiv:1611.01578}, 2016.

\end{thebibliography}
}

\end{document}